\documentclass{article}

\usepackage[final,nonatbib,nonips]{template}

\usepackage[utf8]{inputenc} 
\usepackage[T1]{fontenc}    
\usepackage{hyperref}       
\usepackage{url}            
\usepackage{booktabs}       
\usepackage{amsfonts}       
\usepackage{amsmath}
\usepackage{amssymb}
\usepackage[mathscr]{euscript}
\usepackage{nicefrac}       
\usepackage{microtype}      
\usepackage{tikz,tikz-3dplot}
\usetikzlibrary{shapes,topaths,backgrounds,calc,positioning,fit}
\usepackage{pgfplots}
\pgfplotsset{compat=newest}
\usepackage{floatrow}

\newcommand{\graph}{G}
\newcommand{\espace}{\mathbb{E}}
\newcommand{\nei}{\mathcal{V}}

\title{Generalizing the Convolution Operator\\ to extend CNNs to Irregular Domains}

\author{
  Jean-Charles Vialatte\textsuperscript{1,2}, Vincent Gripon\textsuperscript{2}, Grégoire Mercier\textsuperscript{2} \\ \\
  \begin{tabular}{cc}
  \centering
  \textsuperscript{1}Cityzen Data & \textsuperscript{2}Telecom Bretagne \tabularnewline
  Guipavas, Brittany, France & Brest, Brittany, France 
  \end{tabular} \\ \\
  \textsuperscript{1}\texttt{jean-charles.vialatte@cityzendata.com} \\
  \textsuperscript{2}\texttt{\{jc.vialatte, vincent.gripon, gregoire.mercier\}@telecom-bretagne.eu}
}

\begin{document}

\maketitle

\begin{abstract}
  Convolutional Neural Networks (CNNs) have become the state-of-the-art in supervised learning vision tasks. Their convolutional filters are of paramount importance for they allow to learn patterns while disregarding their locations in input images. When facing highly irregular domains, generalized convolutional operators based on an underlying graph structure have been proposed. However, these operators do not exactly match standard ones on grid graphs, and introduce unwanted additional invariance (e.g. with regards to rotations). We propose a novel approach to generalize CNNs to irregular domains using weight sharing and graph-based operators. Using experiments, we show that these models resemble CNNs on regular domains and offer better performance than multilayer perceptrons on distorded ones.
\end{abstract}

\section{Introduction}

CNNs\cite{lecun1998gradient} are state-of-the-art for performing supervised learning on data defined on lattices~\cite{krizhevsky2012imagenet}. Contrary to classical multilayer perceptrons (MLPs)~\cite{hornik1989multilayer}, CNNs take advantage of the underlying structure of the inputs. When facing data defined on irregular domains, CNNs cannot always be directly applied even if the data may present an exploitable underlying structure.

An example of such data are spatio-temporal time series generated by a set of Internet of Things devices. They typically consist of datapoints irregularly spaced out on a Euclidean space. As a result, a graph $\graph$ can be defined where the vertices are the devices and the edges connect neighbouring ones. Other examples include signals on graphs including brain imaging, transport networks, bag of words graphs\dots In all these examples, a signal can generally be seen as a vector in $\mathbb{R}^d$ where $d$ is the order of the graph. As such, each vertex is associated with a coordinate, and the edges weights represent some association between the corresponding vertices.

Disregarding the graph $\graph$, MLPs can be applied on these datasets. We are interested in defining convolutional operators that are able to exploit $\graph$ as well as a possible embedding Euclidean space. Our motivation is to imitate the gain of performance allowed by CNNs over MLPs on irregular domains.

In graph signal processing, extended convolutional operators using spectral graph theory~\cite{chung1996spectral} have been proposed~\cite{shuman2013emerging}. These operators have been applied to deep learning~\cite{bruna2013spectral} and obtain performance similar to CNNs on regular domains, despite the fact they differ from classical convolutions. However, for slightly distorted domains, the obtained operators become non-localized, thus failing taking account of the underlying structure.

We propose another approach that generalizes convolutional operators by taking into account $\graph$. Namely, we make sure that the proposed solution has properties inherent to convolutions: linearity, locality and kernel weight sharing. We then apply it directly to the graph structure. We use it as a substitution of a convolutional layer in a CNN and stress our resulting technique on comparative benchmarks, showing a significant improvement compared with a MLP. The obtained operator happens to exactly be the classical convolutional one when applied on regular domains.

The outline of the paper is as follows. In Section~\ref{relatedwork}, we discuss related work. In Section~\ref{operator} we introduce our proposed operator. In Section~\ref{cnn} we explain how to apply it to CNN-like structures. Section~\ref{experiments} contains the experiments. Section~\ref{conclusion} is a conclusion.

\section{Related Works}
\label{relatedwork}

  For graph-structured data, Bruna et al.~\cite{bruna2013spectral} have proposed an extension of the CNN using graph signal processing theory~\cite{shuman2013emerging}. Their convolution is defined in the spectral domain related to the Laplacian matrix of the graph. As such, in the case where the graph is a lattice, the construction is analogous to the regular convolution defined in the Fourier domain. The operation is defined as spectral multipliers obtained by smooth interpolation of a weight kernel, and they explain how it ensures the localization property. In that paper, they also define a construction that creates a multi-resolution structure of a graph, for allowing it to support a deep learning architecture. Henaff et al.~\cite{henaff2015deep} have extended Bruna's spectral network to large scale classification tasks and have proposed both supervised and unsupervided methods to find an underlying graph structure when it isn't already given.

  However, when the graph is irregular, they partly lose the localization property of their convolution is partially lost. As the spectral domain is undirected with respect to the graph domain, some sort of rotation invariances are also introduced. Hence, the results in the graph domain aren't ressembling to those of a convolution. In our case, we want to define a convolution supported locally. Moreover, we want that every input can be defined on a different underlying graph structure, so that the learnt filters can be applied on data embedded in the same space regardless of what structure have supported the convolution during the training.

  These properties are also retained by the convolution defined in the ShapeNet paper\cite{masci2015shapenet}, which defines a convolution for data living on non-euclidean manifolds. I.e their construction does maintain the locality and allow for reusing the learnt filters on other manifolds. However, it requires at least a manifold embedding of the data and a geodesic polar coordinates system. Although being less specific, our proposed method will be a strict generalization of CNN in the sense that CNNs are a special case of it.

  In the case where the data is sparse, Graham~\cite{graham2014spatially} has proposed a framework to implement a spatially sparse CNN efficiently. If the underlying graph structure is embedded into an Euclidean space, the convolution we propose in this paper can be seen as spatially sparse too, in the sense that the data has only non-zero coordinates on the vertices of the graph it is defined on. In our case we want to define a convolution for which the inputs can have values on any point of the embedding space, whereas in the regular case inputs can only have values on vertices of an underlying grid.

\section{Proposed convolution operator}
\label{operator}
  \subsection{Definitions}

Let $S$ be a set of points, such that each one defines a set of neighbourhoods.

An \textit{entry} $e$ of a \textit{dataset} $\mathcal{D}$ is a column vector that can be of any size, for which each dimension represents a \textit{value taken by $e$} at a certain point $u \in S$. $u$ is said to be \textit{activated} by $e$. A point $u$ can be associated to at most one dimension of $e$. If it is the $i$th dimension of $e$, then we denote the value taken by $e$ at $u$ by either $e_u$ or $e_i$. $\mathcal{D}$ is said to be \textit{embedded} in $S$.

We say that two entries $e$ and $e'$ are \textit{homogeneous} if they have the same size and if their dimensions are always associated to the same points.

  \subsection{Formal description of the \textit{generalized} convolution}

Let's denote by $\mathcal{C}$ a generalized convolution operator.We want it to observe the following conditions:
\begin{itemize}
  \item Linearity
  \item Locality
  \item Kernel weight sharing
\end{itemize}

As $\mathcal{C}$ must be linear, then for any entry $e$, there is a matrix $W^e$ such that $\mathcal{C}(e) = W^ee$. Unless the entries are all homogeneous, $W^e$ depends on $e$. For example, in the case of regular convolutions on images, $W$ is a Toeplitz matrix and doesn't depend on $e$.

In order to meet the locality condition, we first want that the coordinates of $\mathcal{C}(e)$ have a local meaning. To this end, we impose that $\mathcal{C}(e)$ lives in the same space as $e$, and that $\mathcal{C}(e)$ and $e$ are homogeneous. Secondly, for each $u$, we want $\mathcal{C}(e)_u$ to be only function of values taken by $e$ at points contained in a certain neighbourhood of $u$. It results that rows of $W^e$ are generally sparse.

Let's attribute to $\mathcal{C}$ a kernel of $n$ weights in the form of a row vector $w = (w_1,w_2,..,w_n)$, and let's define the set of allocation matrices $\mathcal{A}$ as the set of binary matrices that have at most one non-zero coordinate per column. As $\mathcal{C}$ must share its weights across the activated points, then for each row $W^e_i$ of $W^e$, there is an allocation matrix $A^e_i \in \mathcal{A}$ such that $W^e_i = wA^e_i$. To maintain locality, the $j$th column of $A^e_i$ must have a non-zero coordinate if and only if the $i$th and $j$th activated points are in a same neighbourhood.

Let's $A^e$ denotes the block column vector that has the matrices $A^e_i$ for attributes, and let's $\otimes$ denotes the tensor product. Hence, $\mathcal{C}$ is defined by a weight kernel $w$ and an allocation map $e \mapsto A^e$ that maintains locality, such that
\begin{align}
\mathcal{C}(e) = (w \otimes A^e)\ .\ e
\end{align}

  \subsection{The underlying graph that supports the generalized convolution}

As vertices, the set of activated points of an entry $e$ defines a complete oriented graph that we call $\graph^e$. If all the entries are homogeneous, then $\graph^e$ is said to be \textit{static}. In this case, we note $\graph$. Otherwise, there can be one different graph per entry so the kernel of weights can be re-used for new entries defined on different graphs.

Suppose we are given $u \mapsto \mathcal{V}_u$ which maps each $u \in S$ to a neighbourhood $\nei_u$. We then define  $\graph^e_\nei$ as the subgraph of $\graph^e$ such that it contains the edge $(u',u)$ if and only if $u' \in \nei_{u}$. Let $a_\nei: e \mapsto A^e$ be an allocation map such that the $j$th column of $A^e_i$ have a non-zero coordinate if and only if $(e_i,e_j)$ is an edge of $\graph^e_\nei$.

Then the generalized convolution of $e$ by the couple $(w,a_\nei)$ is supported by the underlying graph $\graph^e_\nei$ in the sense that $W^e_\nei = w \otimes a_\nei(e)$ is its adjacency matrix. Note that the underlying graph of a regular convolution is a lattice.

Also note that the family $(\mathcal{V}_u)_{u \in S}$ can be seen as local receptive fields for the generalized convolution, and that the map $a_\nei$ can be seen as if it were distributing the kernel weights into each $\mathcal{V}_u$.\\

  \subsubsection*{Remarks}

The generalized convolution has been defined here as an operation between a kernel weight and a vector. If $\graph=(E,V)$ denotes a graph, then such operator $\ast$ with respect to a third rank tensor $A$ could also have been defined as a bilinear operator between two graph signals $f, g \in \mathbb{R}^E$:
\begin{align}
f \ast g &= (f^\top \otimes A)\ .\ g\\
\forall v \in V,\ (f \ast g)(v) &= f^\top A_v\ g, \textit{ with } A_v \in \mathcal{A}
\end{align}\\ 
Note that the underlying graph depends on the entry. So the learnt filter $w$ is reusable regardless of the entry's underlying graph structure. This is not the case for convolutions defined on fixed graphs like in~\cite{bruna2013spectral,henaff2015deep}.

  \subsection{Example of a generalized convolution shaped by a rectangular window}
\label{rectangle}

Let $\espace$ be a two-dimensional Euclidean space and let's suppose here that $S = \espace$.

Let's denote by $\mathcal{C_\mathcal{R}}$, a generalized convolution shaped by a rectangular window $\mathcal{R}$. We suppose that its weight kernel $w$ is of size $N_w = (2p+1)(2q+1)$ and that $\mathcal{R}$ is of width $(2p+1)\mu$ and height $(2q+1)\mu$, where $\mu$ is a given unit scale.

Let $\graph^e_{p,q}$ be the subgraph of $\graph^e$ such that it contains the edge $(u',u)$ if and only if $|u'_x-u_x| \leq (p + \frac{1}2)\mu$ and $|u'_y-u_y| \leq (q + \frac{1}2)\mu$. In other terms, $\graph^e_{p,q}$ connects a vertex $u$ to every vertex contained in $\mathcal{R}$ when centered on $u$.

Then, we define $\mathcal{C_\mathcal{R}}$ as being supported by $\graph^e_{p,q}$. As such, its adjacency matrix acts as the convolution operator. At this point, we still need to affect the kernel weights to its non-zero coordinates, via the edges of $\graph^e_{p,q}$. This amounts for defining explicitely the map $e \mapsto A^e$ for $\mathcal{C_\mathcal{R}}$. To this end, let's consider the grid of same size as that rectangle, which breaks it down into $(2p+1)(2q+1)$ squares of side length $\mu$, and let's associate a different weight to each square. Then, for each edge $(u',u)$, we affect to it the weight associated with the square within which $u'$ falls when the grid is centered on $u$. This procedure allows for the weights to be shared across the edges. It is illustrated on figure~\ref{irrDomain}.

Note that if the entries are homogeneous and the activated points are vertices of a regular grid, then the matrix $W$, independant of $e$, is a Toepliz matrix which acts as a regular convolution operator on the entries $e$. In this case, $\mathcal{C_\mathcal{R}}$ is just a regular convolution. For example, this is the case in section~\ref{link}.

\begin{figure}[h]
  \begin{floatrow}
    \ffigbox{
      \begin{tikzpicture}[scale=1.055]
        \pgfmathsetseed{333}
        \tikzstyle{every node}= [circle,fill=black, minimum size= 4pt,inner sep=0pt];
        \foreach \x in {1,2,4}{
          \foreach \y in {1,2,3,4}{
            \node(\x\y) at (\x+0.5*rand,\y+0.5*rand) {};
          }
        }
        \foreach \y in {1,2,4}{
            \node(3\y) at (3+0.5*rand,3+0.5*rand) {};
        }
        \node(33)[circle,fill=black] at (2.5,2.5) {};
        \foreach \x in {1.6,2.2,2.8,3.4}{
          \path[dashed]
          (\x,1.6) edge (\x,3.4)
          (1.6,\x) edge (3.4,\x);
        }

        \draw[ultra thin] (1.6,1.45) edge[<->] (3.4,1.45);
        \draw[ultra thin] (2.8,3.55) edge[<->] (3.4,3.55);
        \node[draw=none,fill=none] at (2.5,1.25) {\scriptsize{$(2p+1)\mu$}};
        \node[draw=none,fill=none] at (3.15,3.7) {\scriptsize{$\mu$}};

        \draw (0,0) rectangle (5,5);
        \node[draw=none,fill=none] at (0.7,0.3) {$S=\espace$};
      \end{tikzpicture}
    }
    {
      \caption{Example of a moving grid. The grid defines the allocation of the kernel weights.}
      \label{irrDomain}
    }
    \ffigbox{
      \begin{tikzpicture}[scale=0.75]
        \tikzstyle{every node}= [circle,fill=black, minimum size= 3pt,inner sep=0pt];
          \foreach \x in {1,2,3,4,5,6}{
            \foreach \y in {1,2,3,4,5,6}{
              \node(\x\y) at (\x,\y) {};
            }
          }

          \foreach \x in {1.5,2.5,3.5,4.5}{
            \path[thin]
            (\x,1.5) edge (\x,4.5)
            (1.5,\x) edge (4.5,\x);
          }

          \draw[ultra thin] (1,6.17) edge[<->] (2,6.17);
          \node[draw=none,fill=none] at (1.5,6.3) {\scriptsize{$\mu$}};

          \foreach \x in {1,2,3,4,5,6}{
            \draw[dotted,thin] (1,\x) -- (6,\x);
            \draw[dotted,thin] (\x,1) -- (\x,6);
          }

          \foreach \x in {1,2,3,4,5}{
            \draw[dotted,thin] (1,\x) -- (7-\x,6);
            \draw[dotted,thin] (6,\x) -- (\x,6);
          }

          \foreach \x in {2,3,4,5}{
            \draw[dotted,thin] (\x,1) -- (6,7-\x);
            \draw[dotted,thin] (7-\x,1) -- (1,7-\x);
          }

          \draw (0,0) rectangle (7,7);

          \node[draw=none,fill=none] at (0.9,0.4) {$S=\espace$};
      \end{tikzpicture}
    }
    {
      \caption{On regular domains, the moving grid allocates the weights similarly to the moving window of a standard convolution.}
      \label{regDomain}
    }
  \end{floatrow}
\end{figure}

  \subsection{Link with the standard convolution on image datasets}
  \label{link}

  Let $\mathcal{D}$ be an image dataset. Entries of $\mathcal{D}$ are homogeneous and their dimensions represent the value at each pixel. In this case, we can set $S=\espace$, of dimension 2, such that each pixel is located at entire coordinates. More precisely, if the images are of width $n$ and height $m$, then the pixels are located at coordinates $(i,j) \in \{0, 1, 2, ..., n\} \text{x} \{0, 1, 2, ..., m\}$. Hence, the pixels lie on a regular grid and thus are spaced out by a constant distance $\mu = 1$.

  Let's consider the static underlying graph $\graph_{p,q}$ and the generalized convolution by a rectangular window $\mathcal{C_\mathcal{R}}$, as defined in the former section. Then, applying the same weight allocation strategy will lead to affect every weight of the kernel into the moving window $\mathcal{R}$. Except on the border, one and only one pixel will fall into each square of the moving grid at each position, as depicted in figure~\ref{regDomain}. Indeed, $\mathcal{R}$ behaves exactly like a moving window of the standard convolution, except that it considers that the images are padded with zeroes on the borders.

  \section{Application to CNNs}
\label{cnn}

  \subsection{Neural network interpretation}

Let $\mathcal{L}^d$ and $\mathcal{L}^{d+1}$ be two layers of neurons, such that forward-propagation is defined from $\mathcal{L}^d$ to $\mathcal{L}^{d+1}$. Let's define such layers as a set of neurons being located in $S$. These layers must contain as many neurons as points that can be activated. In other terms, $S \cong \mathcal{L}^d \cong \mathcal{L}^{d+1}$. As such, we will abusively use the term of \textit{neuron} instead of \textit{point}.

  \begin{figure}[h]
    \centering
    \includegraphics{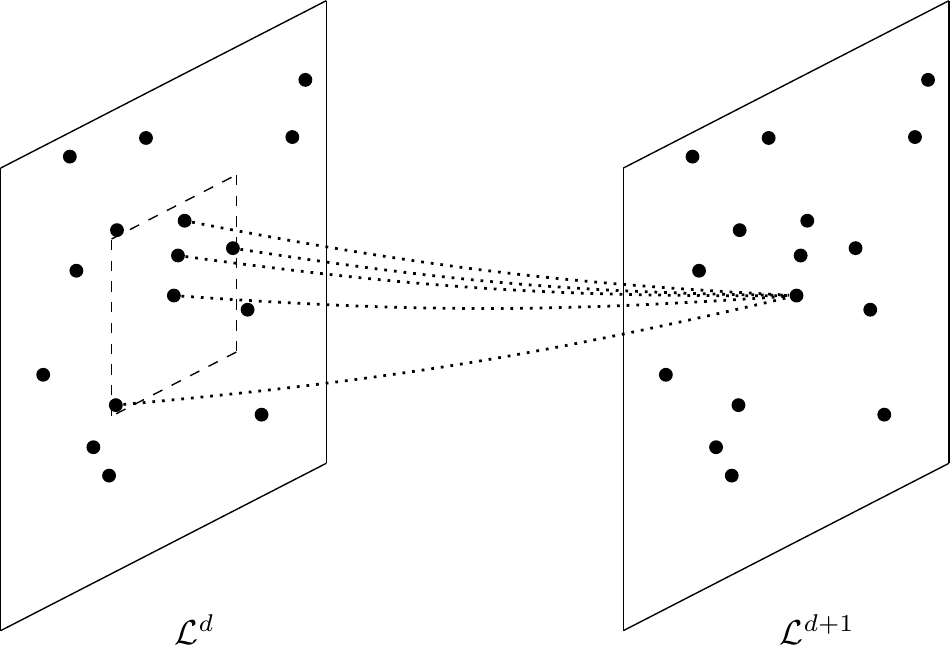}
    \caption{Generalized convolution between two layers of neurons.}
    \label{nn}
  \end{figure}

The \textit{generalized} convolution between these two layers can be interpreted as follow. An entry $e$ activates the same $N$ neurons in each layer. Then, a convolution shape takes $N$ positions onto $\mathcal{L}^d$, each position being associated to one of the activated neurons of $\mathcal{L}^{d+1}$. At each position, connections are drawn from the activated neurons located inside the convolution shape in destination to the associated neuron. And a subset of weights from $w$ are affected to these connections, according to a weight sharing strategy defined by an allocation map. Figure~\ref{nn} illustrates a convolution shaped by a rectangular window.

The forward and backward propagation between $\mathcal{L}^d$ and $\mathcal{L}^{d+1}$  are applied using the described neurons and connections. After a generalized convolution operation, an activation function is applied on the output neurons. Then a pooling operation is done spatially: the input layer is divided into patches of same size, and all activated neurons included in this patch are pooled together. Unlike a standard pooling operation, the number of activated neurons in a patch may vary.

Generalized convolution layers can be vectorized. They can have multiple input channels and multiple feature maps. They shall naturally be placed into the same kind of deep neural network structure than in a CNN. Thus, they are for the irregular input spaces what the standard convolution layers are for regular input spaces.

  \subsection{Implementation}

There are two main strategies to implement the propagations. The first one is to start from (1), derive it and vectorize it. It implies handling semi-sparse representations to minimize memory consumption and to use adequate semi-sparse tensor products.

Instead, we decide to use the neural network interpretation and the underlying graph structure whose edges amount for neurons connections. By this mean, the sparse part of the computations is included via the use of this graph. Also, computations on each edge can be parallelized.

  \subsection{Forward and back propagation formulae}
  \label{formulae}

Let's first recall the propagation formulae from a neural network point of view.

Let's denote by $e_u$ the value of a neuron of $\mathcal{L}^d$ located at $u \in S$, by $f_v$ for a neuron of $\mathcal{L}^{d+1}$, and by $g_v$ if this neuron is activated by the activation function $\sigma$.

We denote by $\textit{prev}(v)$ the set neurons from the previous layer connected to $v$, and by $\textit{next}(u)$ those of the next layer connected to $u$. $\textit{w}_{uv}$ is the weight affected to the connection between neurons $u$ and $v$. $b$ is the bias term associated to $\mathcal{L}^{d+1}$.

After the forward propagation, values of neurons of $\mathcal{L}^{d+1}$ are determined by those of $\mathcal{L}^d$:
\begin{align}
f_v &= \sum_{u \in \textit{prev}(v)}{e_u\textit{w}_{uv}} \\
g_v &= \sigma(f_v + b)
\end{align}

Thanks to the chain rule, we can express derivatives of a layer with those of the next layer:
\begin{align}
\delta_v = \frac{\partial{E}}{\partial{f_v}} &= \frac{\partial{E}}{\partial{g_v}}\frac{\partial{g_v}}{\partial{f_v}} = \frac{\partial{E}}{\partial{g_v}}\sigma^{'}(f_v + b) \\
\frac{\partial{E}}{\partial{e_u}} &= \sum_{v \in \textit{next(u)}}{\frac{\partial{E}}{\partial{f_v}}\frac{\partial{f_v}}{\partial{e_u}}} = \sum_{v \in \textit{next(u)}}{\delta_v\textit{w}_{uv}}
\end{align}

We call $\textit{edges}(\textit{w})$ the set of edges to which the weight $\textit{w}$ is affected. If $\omega \in \textit{edges}(\textit{w})$, $f_{\omega^+}$ denotes the value of the destination neuron, and $e_{\omega^-}$ denotes the value of the origin neuron.

The back propagation allows to express the derivative of any weight $\textit{w}$:
\begin{align}
\frac{\partial{E}}{\partial{\textit{w}}} &= \sum_{\omega \in \textit{edges}(\textit{w})}{\frac{\partial{E}}{\partial{f_{\omega^+}}}\frac{\partial{f_{\omega^+}}}{\partial{\textit{w}}}} = \sum_{\omega \in \textit{edges}(\textit{w})}{\delta_{\omega^+}e_{\omega^-}} \\
\frac{\partial{E}}{\partial{b}} &= \sum_{v}{\frac{\partial{E}}{\partial{g_v}}\frac{\partial{g_v}}{\partial{b}}} = \sum_{v}{\delta_v}
\end{align}

The sets $\textit{prev}(v)$, $\textit{next}(u)$ and $\textit{edges}(\textit{w})$ are determined by the graph structure, which in turn is determined beforehand by a procedure like the one described in section~\ref{rectangle}. The particularization of the propagation formulae with these sets is the main difference with the standard formlulae.

  \subsection{Vectorization}

Computations are done per batch of entries $B$. Hence, the graph structure used for the computations must contain the weighted edges of all entries $e \in B$. If necessary, entries of $B$ are made homogeneous: if a neuron $u$ is not activated by an entry $e$ but is activated by another entry of $B$, then $e_u$ is defined and is set to zero.

The third-rank tensor counterparts of $e$, $f$, and $g$ are thus denoted by $\mathscr{E}$, $\mathscr{F}$ and $\mathscr{G}$. Their third dimension indexes the channels (input channels or feature maps). Their submatrix along the neuron located at $x \in S$ are denoted $\mathscr{E}_x$, $\mathscr{F}_x$ and $\mathscr{G}_x$, rows are indexing entries and columns are indexing channels.
The counterparts of $\textit{w}$ and $b$ are $\mathscr{W}$ and $\beta$. The first being a third-rank tensor and the second being a vector with one value per feature map. $\widetilde{\beta}$ denotes the 3D tensor obtained by broadcasting $\beta$ along the two other dimensions of $\mathscr{F}$. $\mathscr{W}_\textit{w}$ denotes the submatrix of $\mathscr{W}$ along the kernel weight $\textit{w}$. Its rows index the feature maps and its columns index the input channels.

With these notations, the convolution formulae (1) rewrites:
\begin{align}
\mathcal{C}(\mathscr{E}) = (\mathscr{W} \otimes A^\mathscr{E})\ \boxtimes\ \mathscr{E}
\end{align}
$\otimes$ and $\boxtimes$ being tensor products. $\otimes$ is contracted along the dimensions that index the kernel weights, and $\boxtimes$ is contracted along the dimensions that index the points present in the entries.

The vectorized counterparts of the formulae from section~\ref{formulae} can be obtained in the same way:
\begin{align}
\mathscr{F}_v &= \sum_{u \in \textit{prev}(v)}{\mathscr{E}_u\mathscr{W}_{\textit{w}_{uv}}^\top} \\
\mathscr{G} &= \sigma(\mathscr{F} + \widetilde{\beta}) \\
\Delta = \left(\frac{\partial{E}}{\partial{\mathscr{F}}}\right) &= \left(\frac{\partial{E}}{\partial{\mathscr{G}}}\right)\circ\ \sigma^{'}(\mathscr{F} + \widetilde{\beta}) \\
\left(\frac{\partial{E}}{\partial{\mathscr{E}}}\right)_u &= \sum_{v \in \textit{next(u)}}{\Delta_v\mathscr{W}_{uv}} \\
\left(\frac{\partial{E}}{\partial{\mathscr{W}}}\right)_\textit{w} &= \sum_{\omega \in \textit{edges}(\textit{w})}{\Delta_{\omega^+}^\top\mathscr{E}_{\omega^-}} \\
\frac{\partial{E}}{\partial{\beta}} &= \sum_{j}{\left(\sum_{v}{\Delta_v}\right)_{j\textit{-th column}}}
\end{align}

Where $\circ$ and $^\top$ respectively denotes Hadamard product and transpose.

\section{Experiments}
\label{experiments}

In order to measure the gain of performance allowed by the generalized CNN over MLP  on irregular domains, we made a series of benchmarks on distorded versions of the MNIST dataset~\cite{lecun1998mnist}, consisting of images of 28x28 pixels. To distort the input domain, we plunged the images into a 2-d euclidean space by giving entire coordinates to pixels. Then, we applied a gaussian displacement on each pixel, thus making the data irregular and unsuitable for regular convolutions. For multiple values of standard deviation of the displacement, we trained a generalized CNN and compared it with a MLP that has the same number of parameters. We choosed a simple yet standard architecture in order to better see the impact of generalized layers.

The architecture used is the following: a generalized convolution layer with relu~\cite{glorot2011deep} and max pooling, made of 20 feature maps, followed by a dense layer and a softmax output layer. The generalized convolution is shaped by a rectangular window of width and height $5\mu$ where the unit scale $\mu$ is chosen to be equal to original distance between two pixels. The max-pooling is done with square patches of side length $2\mu$. The dense layer is composed of 500 hidden units and is terminated by relu activation as well. In order to have the same number of parameters, the compared MLP have 2 dense layers of 500 hidden units each and is followed by the same output layer. For training, we used stochastic gradient descent~\cite{bottou2010large} with Nesterov momentum~\cite{sutskever2013importance} and a bit of L2 regularization~\cite{ng2004feature}.

\begin{figure}[!h]
  \centering
  \includegraphics{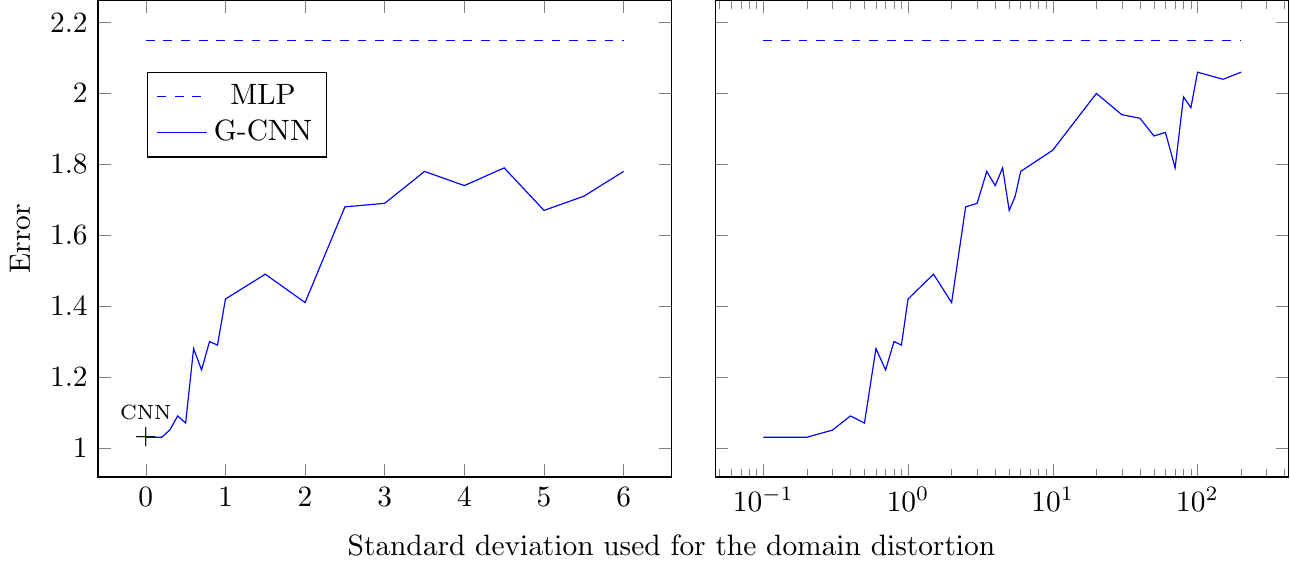}
  \caption{Results for Generalized CNN and MLP}
  \label{distordedDomain}
\end{figure}

 The plot drawn on figure~\ref{distordedDomain} illustrates the gain of performance of a generalized convolutional layer over a dense layer with equal number of parameters. After 15 epochs for both, it shows that the generalized CNN on a distorded domain performs better than the MLP. Indeed, in the case of no domain distortion, the score is the same than a CNN with zero-padding. The error rate goes up a bit with distortion. But even at $200\mu$, the generalized CNN still performs better than the MLP.

  \section{Conclusion and future work}
  \label{conclusion}

  In this paper, we have defined a generalized convolution operator. This operator makes possible to transport the CNN paradigm to irregular domains. It retains the proprieties of a regular convolutional operator. Namely, it is linear, supported locally and uses the same kernel of weights for each local operation. The generalized convolution operator can then naturally be used instead of convolutional layers in a deep learning framework. Typically, the created model is well suited for input data that has an underlying graph structure.

  The definition of this operator is flexible enough for it allows to adapt its weight-allocation map to any input domain, so that depending on the case, the distribution of the kernel weight can be done in a way that is natural for this domain. However, in some cases, there is no natural way but multiple acceptable methods to define the weight allocation. In further works, we plan to study these methods. We also plan to apply the generalized operator on unsupervised learning tasks.

\subsubsection*{Acknowledgements}

I would like to thank my academic mentors, Vincent Gripon and Grégoire Mercier who helped me in this work, as well as my industrial mentor, Mathias Herberts who gave me insights in view of applying the designed model to industrial datasets. \\
This work was partly funded by \href{http://www.cityzendata.com/}{Cityzen Data}, the company behind the \href{http://www.warp10.io}{Warp10} platform, and by the \href{http://www.anrt.asso.fr/fr/espace_cifre/accueil.jsp}{ANRT} (Agence Nationale de la Recherche et de la Technologie) through a CIFRE (Convention Industrielle de Formation par la REcherche), and also by the European Research Council under the European Union's Seventh Framework Program (FP7/2007-2013) / ERC grant agreement number 290901.

\bibliographystyle{ieeetr}
\bibliography{arxiv5}

\end{document}